\documentclass[11pt,a4paper]{article}
\usepackage[hyperref]{acl2018}
\usepackage{times}
\usepackage{latexsym}

\usepackage{url}
\usepackage{amssymb,mathtools}
\usepackage{subfig}
\usepackage{enumitem}

\aclfinalcopy 
\def\aclpaperid{12} 

\title{Polarity and Intensity: the Two Aspects of Sentiment Analysis}

\author{Leimin Tian \\
  School of Informatics \\
  the University of Edinburgh \\
  {\tt Leimin.Tian@monash.edu} \\\And
  Catherine Lai \\
  School of Informatics \\
  the University of Edinburgh \\
  {\tt clai@inf.ed.ac.uk} \\\And
  Johanna D. Moore \\
  School of Informatics \\
  the University of Edinburgh \\
  {\tt J.Moore@ed.ac.uk} \\
}

\date{}

\begin{document}
\maketitle
\begin{abstract}
Current multimodal sentiment analysis frames sentiment score prediction as a general Machine Learning task. However, what the sentiment score actually represents has often been overlooked. As a measurement of opinions and affective states, a sentiment score generally consists of two aspects: polarity and intensity. We decompose sentiment scores into these two aspects and study how they are conveyed through individual modalities and combined multimodal models in a naturalistic monologue setting. In particular, we build unimodal and multimodal multi-task learning models with sentiment score prediction as the main task and polarity and/or intensity classification as the auxiliary tasks. Our experiments show that sentiment analysis benefits from multi-task learning, and individual modalities differ when conveying the polarity and intensity aspects of sentiment.
\end{abstract}

\section{Introduction}\label{sec_intro}
Computational analysis of human multimodal language is a growing research area in Natural Language Processing (NLP). One important type of information communicated through human multimodal language is sentiment. Current NLP studies often define sentiments using scores on a scale, e.g., a 5-point Likert scale representing sentiments from strongly negative to strongly positive. Previous work on multimodal sentiment analysis has focused on identifying effective approaches for sentiment score prediction (e.g.,~\citet{cmumoseiacl2018}). However, in these studies sentiment score prediction is typically represented as a regression or classification task, without taking into account what the sentiment score means. As a measurement of human opinions and affective states, a sentiment score can often be decomposed into two aspects: the polarity and intensity of the sentiment. In this work, we study how individual modalities and multimodal information convey these two aspects of sentiment. 

More specifically, we conduct experiments on the Carnegie Mellon University Multimodal Opinion Sentiment Intensity (CMU-MOSI) database \citep{zadeh2016multimodal}. The CMU-MOSI database is a widely used benchmark database for multimodal sentiment analysis. It contains naturalistic monologues expressing opinions on various subjects. Sentiments are annotated as continuous scores for each opinion segment in the CMU-MOSI database, and data were collected over the vocal, visual, and verbal modalities. We build unimodal and multimodal multi-task learning models with sentiment score regression as the main task, and polarity and/or intensity classification as the auxiliary tasks. Our main research questions are:
\begin{enumerate}[noitemsep,topsep=1pt,leftmargin=*]
\item Does sentiment score prediction benefit from multi-task learning?
\item Do individual modalities convey the polarity and intensity of sentiment differently?
\item Does multi-task learning influence unimodal and multimodal sentiment analysis models in different ways?
\end{enumerate}

Our work contributes to our current understanding of the intra-modal and inter-modal dynamics of how sentiments are communicated in human multimodal language. Moreover, our study provides detailed analysis on how multi-task learning and modality fusion influences sentiment analysis.

\section{Background}\label{sec_bg}
Sentiment is an important type of information conveyed in human language. Previous sentiment analysis studies in the field of NLP have mostly been focused on the verbal modality (i.e., text). For example, predicting the sentiment of Twitter texts \citep{kouloumpis2011twitter} or news articles \citep{balahur2013sentiment}. However, human language is multimodal in, for instance, face-to-face communication and online multimedia opinion sharing. Understanding natural language used in such scenarios is especially important for NLP applications in Human-Computer/Robot Interaction. Thus, in recent years there has been growing interest in multimodal sentiment analysis. The three most widely studied modalities in current multimodal sentiment analysis research are: vocal (e.g., speech acoustics), visual (e.g., facial expressions), and verbal (e.g., lexical content). These are sometimes referred to as ``the three Vs'' of communication \citep{mehrabian1971silent}. Multimodal sentiment analysis research focuses on understanding how an individual modality conveys sentiment information (intra-modal dynamics), and how they interact with each other (inter-modal dynamics). It is a challenging research area and state-of-the-art performance of automatic sentiment prediction has room for improvement compared to human performance \citep{zadeh2018multi}.

While multimodal approaches to sentiment analysis are relatively new in NLP, multimodal emotion recognition has long been a focus of Affective Computing. For example, \citet{de2000bimodal} combined facial expressions and speech acoustics to predict the Big-6 emotion categories \citep{big-6}. Emotions and sentiments are closely related concepts in Psychology and Cognitive Science research, and are often used interchangeably. \citet{munezero2014they} identified the main differences between sentiments and emotions to be that sentiments are more stable and dispositional than emotions, and sentiments are formed and directed toward a specific object. However, when adopting the cognitive definition of emotions which connects emotions to stimuli in the environment \citep{ortony1990cognitive}, the boundary between emotions and sentiments blurs. In particular, the circumplex model of emotions proposed by \citet{russell1980circumplex} describes emotions with two dimensions: Arousal which represents the level of excitement (active/inactive), and Valence which represents the level of liking (positive/negative). In many sentiment analysis studies, sentiments are defined using Likert scales with varying numbers of steps. For example, the Stanford Sentiment Treebank \citep{socher2013recursive} used a 7-point Likert scale to annotate sentiments. Such sentiment annotation schemes have two aspects: polarity (positive/negative values) and intensity (steps within the positive or negative range of values). This similarity suggests connections between emotions defined in terms of Valence and Arousal, and sentiments defined with polarity and intensity, as shown in Table~\ref{bg_emo}. However, while previous work on multimodal emotion recognition often predicts Arousal and Valence separately, most previous work on multimodal sentiment analysis generally predicts the sentiment score as a single number. Thus, we are motivated to study how the polarity and intensity aspects of sentiments are each conveyed.

\begin{table}[!htb]
\begin{center}
\begin{tabular}{|l|c|c|}
\hline
Aspect of the affect & Activeness & Liking \\
\hline
Emotion as by & Arousal & Valence \\
\citet{russell1980circumplex} & & \\
\hline
Sentiment on & Intensity & Polarity \\
a Likert scale & & \\
\hline
\end{tabular}
\end{center}
\caption{Similarity between circumplex model of emotion and Likert scale based sentiment.}
\label{bg_emo}
\end{table}

In order to decompose sentiment scores into polarity and intensity and study how they are conveyed through different modalities, we include polarity and/or intensity classification as auxiliary tasks to sentiment score prediction with multi-task learning. One problem with Machine Learning approaches for Affective Computing is model robustness. In multi-task learning, the model shares representations between the main task and auxiliary tasks related to the main task, often enabling the model to generalize better on the main task \citep{ruder2017overview}. Multiple auxiliary tasks have been used in previous sentiment analysis and emotion recognition studies. For example, \citet{xia2017multi} used dimensional emotion regression as an auxiliary task for categorical emotion classification, while \citet{chen2017improving} used sentence type classification (number of opinion targets expressed in a sentence) as an auxiliary task for verbal sentiment analysis. To the best of our knowledge, there has been no previous work applying multi-task learning to the CMU-MOSI database.

In addition to how individual modalities convey sentiment, another interesting topic in multimodal sentiment analysis is how to combine information from multiple modalities. There are three main types of modality fusion strategies in current multimodal Machine Learning research \citep{baltruvsaitis2018multimodal}: early fusion which combines features from different modalities, late fusion which combines outputs of unimodal models, and hybrid fusion which exploits the advantages of both early and late fusion. We will study the performance of these different modality fusion strategies for multimodal sentiment analysis.

\section{Methodology}\label{sec_metho}

\subsection{The CMU-MOSI Database}\label{sec_mosi}
The CMU-MOSI database contains 93 YouTube opinion videos from 89 distinct speakers \citep{zadeh2016multimodal}. The videos are monologues on various topics recorded with various setups, lasting from 2 to 5 minutes. 2199 opinion segments were manually identified from the videos with an average length of 4.2 seconds (approximately 154 minutes in total). An opinion segment is the expression of opinion on a distinct subject, and can be part of a spoken utterance or consist of several consecutive utterances. \citet{zadeh2016multimodal} collected sentiment score annotations of the opinion segments using Amazon Mechanical Turk and each video clip was annotated by five workers. For each opinion segment the sentiment scores are annotated on a 7-point Likert scale, i.e., strongly negative (\texttt{-}3), negative (\texttt{-}2), weakly negative (\texttt{-}1), neutral (0), weakly positive (\texttt{+}1), positive (\texttt{+}2), strongly positive (\texttt{+}3). The gold-standard sentiment score annotations provided are the average of all five workers.

Previous work on the CMU-MOSI database explored various approaches to improving performance of sentiment score prediction (e.g.,~\citet{cmumoseiacl2018}). The target sentiment annotations can be continuous sentiment scores or discrete sentiment classes (binary, 5-class, or 7-class sentiment classes). The Tensor Fusion Network model of \citet{zadeh2017tensor} achieved the best performance for continuous sentiment score regression on the CMU-MOSI database using features from all three modalities. The Pearson's correlation coefficient between the automatic predictions of their model and the gold-standard sentiment score annotations reached 0.70. In this work, we follow the parameter settings and features used by \citet{zadeh2017tensor} when predicting the sentiment scores.

\subsection{Multimodal Sentiment Analysis with Multi-Task Learning}\label{sec_model}
In this study, we apply multi-task learning to sentiment analysis using the CMU-MOSI database. We consider predicting the gold-standard sentiment scores as the main task. Thus, the single-task learning model is a regression model predicting the sentiment score $S_{o}$ of an opinion segment \textit{o}, which has a value within range [\texttt{-}3,\texttt{+}3]. To perform multi-task learning, for each opinion segment, we transform the gold-standard sentiment score $S_{o}$ into binary polarity class $P_{o}$ and intensity class $I_{o}$:
\begin{equation}
P_{o}= 
    \begin{cases}
    \text{Positive},& \text{if } S_{o} \geq 0\\
    \text{Negative},& \text{if } S_{o} < 0
    \end{cases}
\end{equation}

\begin{equation}
I_{o}= 
    \begin{cases}
    \text{Strong},& \text{if } |S_{o}| \geq 2.5\\
    \text{Medium},& \text{if } 1.5 \leq |S_{o}| < 2.5\\
    \text{Weak},& \text{if } 0.5 \leq |S_{o}| < 1.5\\
    \text{Neutral},& \text{if } |S_{o}| < 0.5
    \end{cases}
\end{equation}

Unlike previous studies performing a 5-class or 7-class classification experiment for sentiment analysis, our definition of intensity classes uses the absolute sentiment scores, thus separating the polarity and intensity information. For example, an opinion segment $o_{1}$ with $S_{o_{1}}$ = \texttt{+}3.0 will have $P_{o_{1}}$ = \textit{Positive} and $I_{o_{1}}$ = \textit{Strong}, while an opinion segment $o_{2}$ with $S_{o_{2}}$ = \texttt{-}2.75 will have $P_{o_{2}}$ = \textit{Negative} and $I_{o_{2}}$ = \textit{Strong}. Note that here we group the sentiment scores into discrete intensity classes. In the future we plan to study the gain of preserving the ordinal information between the intensity classes.

For each modality or fusion strategy we build four models: single-task sentiment regression model, bi-task sentiment regression model with polarity classification as the auxiliary task, bi-task sentiment regression model with intensity classification as the auxiliary task, and tri-task sentiment regression model with both polarity and intensity classification as the auxiliary tasks. In the bi-task and tri-task models, the main task loss is assigned a weight of 1.0, while the auxiliary task losses are assigned a weight of 0.5. Structures of the single-task and multi-task learning models only differ at the output layer: for sentiment score regression the output is a single node with tanh activation; for polarity classification the output is a single node with sigmoid activation; for intensity classification the output is 4 nodes with softmax activation. The main task uses mean absolute error as the loss function, while polarity classification uses binary cross-entropy as the loss function, and intensity classification uses categorical cross-entropy as the loss function. Following state-of-the-art on the CMU-MOSI database \citep{zadeh2017tensor}, during training we used Adam as the optimization function with a learning rate of 0.0005. We use the CMU Multimodal Data Software Development Kit (SDK) \citep{zadeh2018multi} to load and pre-process the CMU-MOSI database, which splits the 2199 opinion segments into training (1283 segments), validation (229 segments), and test (686 segments) sets.\footnote{Segment 13 of video 8qrpnFRGt2A is partially missing and thus removed for the experiments.} We implement the sentiment analysis models using the Keras deep learning library \citep{chollet2015keras}.

\subsection{Multimodal Features}\label{sec_feat}
\begin{figure}[!htb]
\centering
\includegraphics[width=0.75\linewidth]{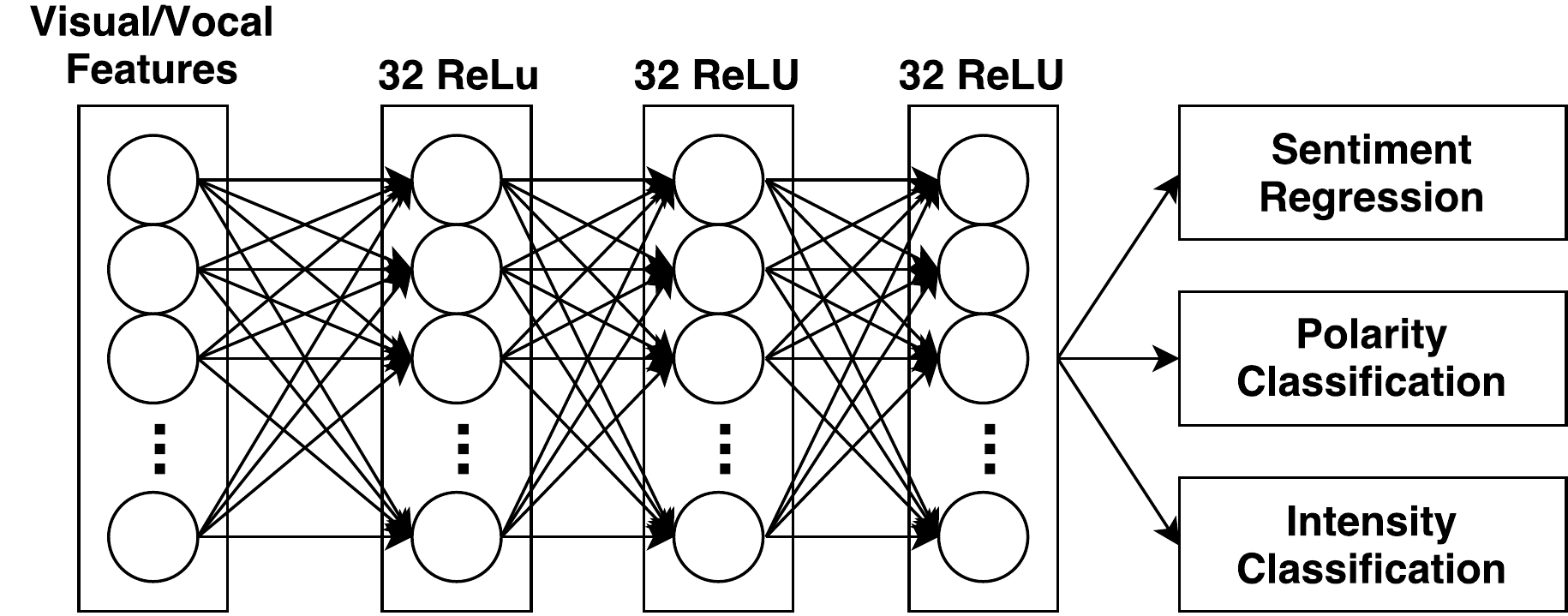}
\caption{Visual/vocal unimodal tri-task model}\label{fig_AV}
\end{figure}

For the vocal modality, we use the COVAREP feature set provided by the SDK. These are 74 vocal features including 12 Mel-frequency cepstral coefficients, pitch tracking and voiced/unvoiced segmenting features, glottal source parameters, peak slope parameters, and maxima dispersion quotients. The vocal features are extracted from the audio recordings at a sampling rate of 100Hz. For the visual modality, we use the FACET feature set provided by the SDK. These are 46 visual features including facial indicators of 9 types of emotion (anger, contempt, disgust, fear, joy, sadness, surprise, frustration, and confusion) and movements of 20 facial action units. The visual features are extracted from the speaker's facial region in the video recordings at a sampling rate of 30Hz. Following \citet{zadeh2017tensor}, for the vocal and visual unimodal models, we apply a drop-out rate of 0.2 to the features and build a neural network with three hidden layers of 32 ReLU activation units, as shown in Figure~\ref{fig_AV}.

\begin{figure}[!htb]
\centering
\includegraphics[width=0.7\linewidth]{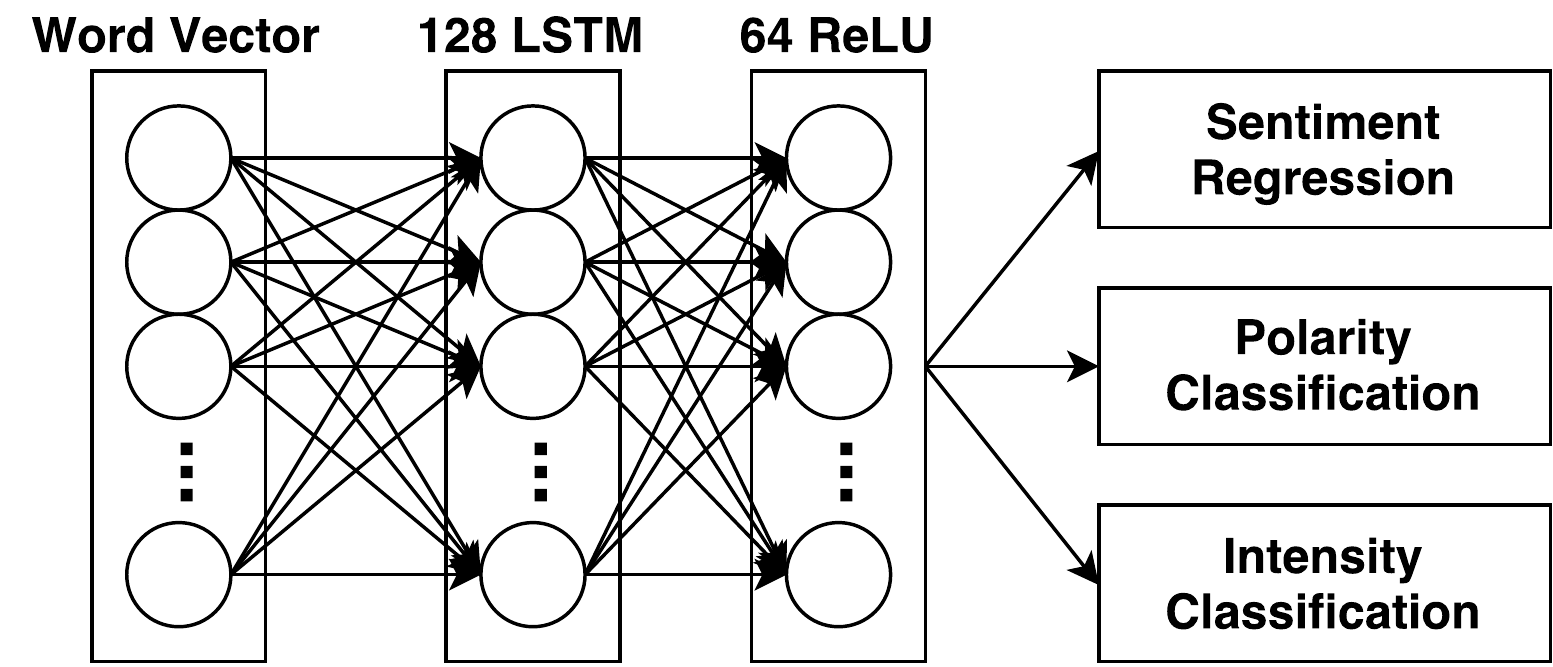}
\caption{Verbal unimodal tri-task model}\label{fig_T}
\end{figure}

For the verbal modality, we use the word embedding features provided by the SDK, which are 300-dimensional GloVe word vectors. There are 26,295 words in total (3,107 unique words) in the opinion segments of the CMU-MOSI database. Following \citet{zadeh2017tensor}, for the verbal unimodal model we build a neural network with one layer of 128 Long Short-Term Memory (LSTM) units and one layer of 64 ReLU activation units, as shown in Figure~\ref{fig_T}. Previous work has found that context information is important for multimodal sentiment analysis, and the use of LSTM allows us to include history \citep{poria2017context}.

Note that the visual and vocal features are extracted at the frame level, while the verbal features are extracted at the word level. Before conducting all unimodal and multimodal experiments, we aligned all the features to the word level using the SDK. This down-samples the visual and vocal features to the word level by computing the averaged feature vectors for all frames within a word.

\subsection{Modality Fusion Strategies}\label{sec_fuse}
We test four fusion strategies here: Early Fusion (EF), Tensor Fusion Network (TFN), Late Fusion (LF), and Hierarchical Fusion (HF). EF and LF are the most widely used fusion strategies in multimodal recognition studies and were shown to be effective for multimodal sentiment analysis \citep{poria2015deep}. TFN achieved state-of-the-art performance on the CMU-MOSI database \citep{zadeh2017tensor}. HF is a form of hybrid fusion strategy shown to be effective for multimodal emotion recognition \citep{tian2016recognizing}.

\begin{figure}[!htb]
\centering
\includegraphics[width=0.75\linewidth]{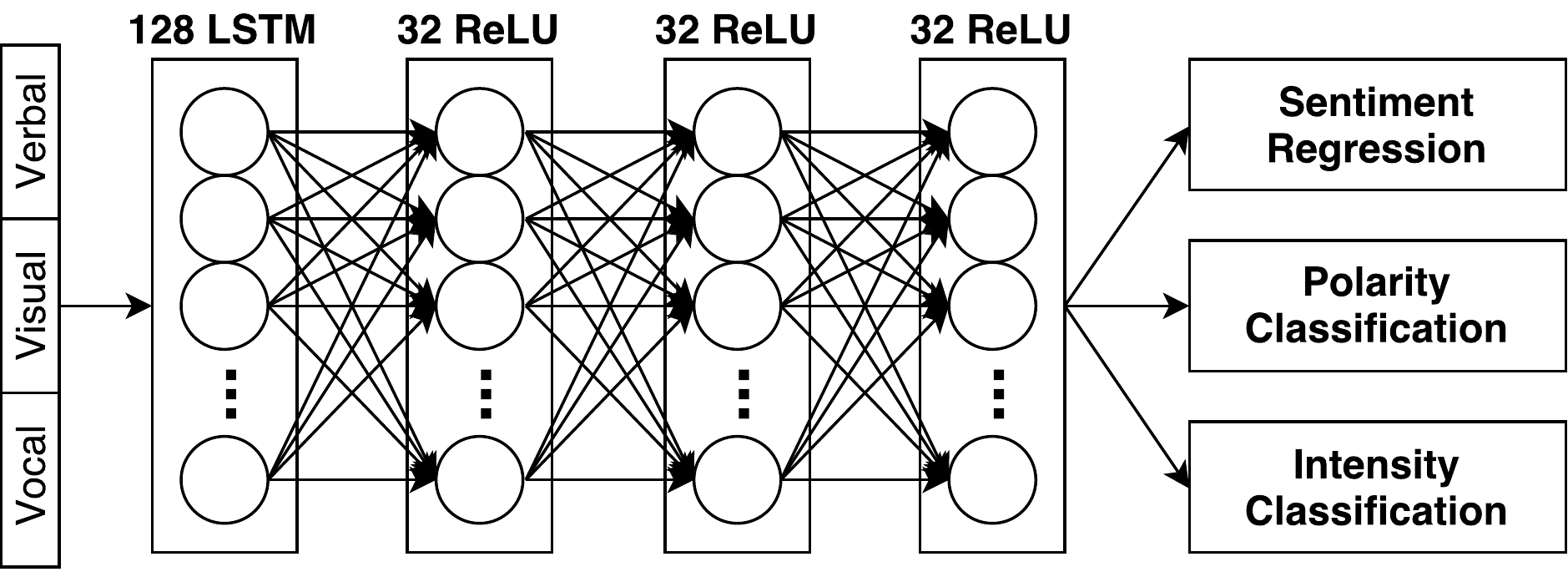}
\caption{Structure of EF tri-task model}\label{fig_FL}
\end{figure}

\begin{figure}[!htb]
\centering
\includegraphics[trim=5.5cm 0 0 0, clip, width=0.6\linewidth]{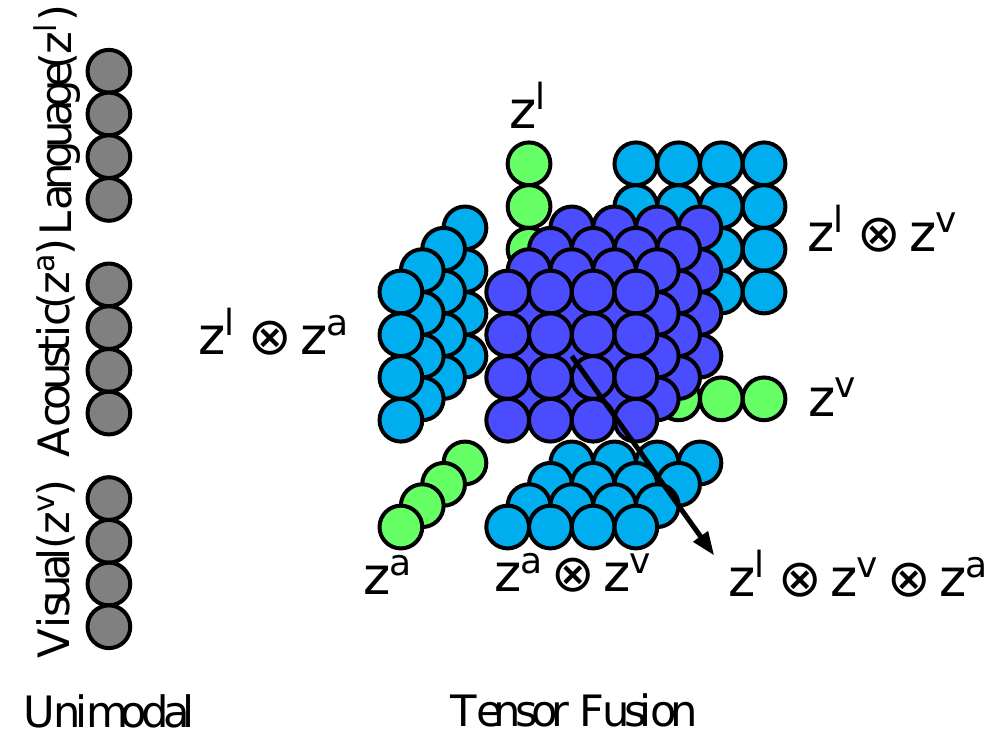}
\caption{Fusion strategy of the TFN model \citep{zadeh2017tensor}}\label{fig_TFN}
\end{figure}

\begin{figure}[!htb]
\centering
\includegraphics[width=0.95\linewidth]{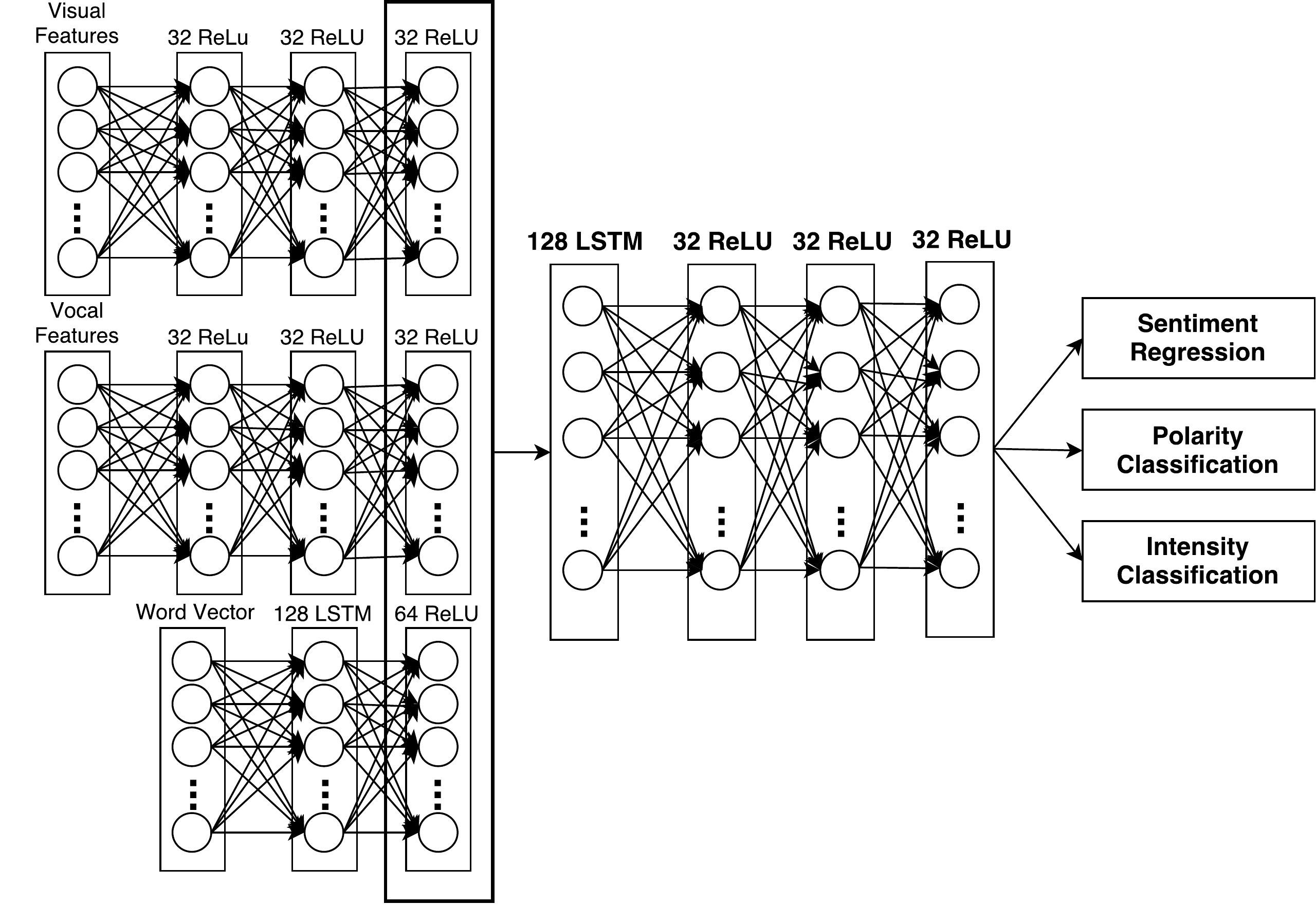}
\caption{Structure of LF tri-task model}\label{fig_DL}
\end{figure}

\begin{figure}[!htb]
\centering
\includegraphics[width=1.0\linewidth]{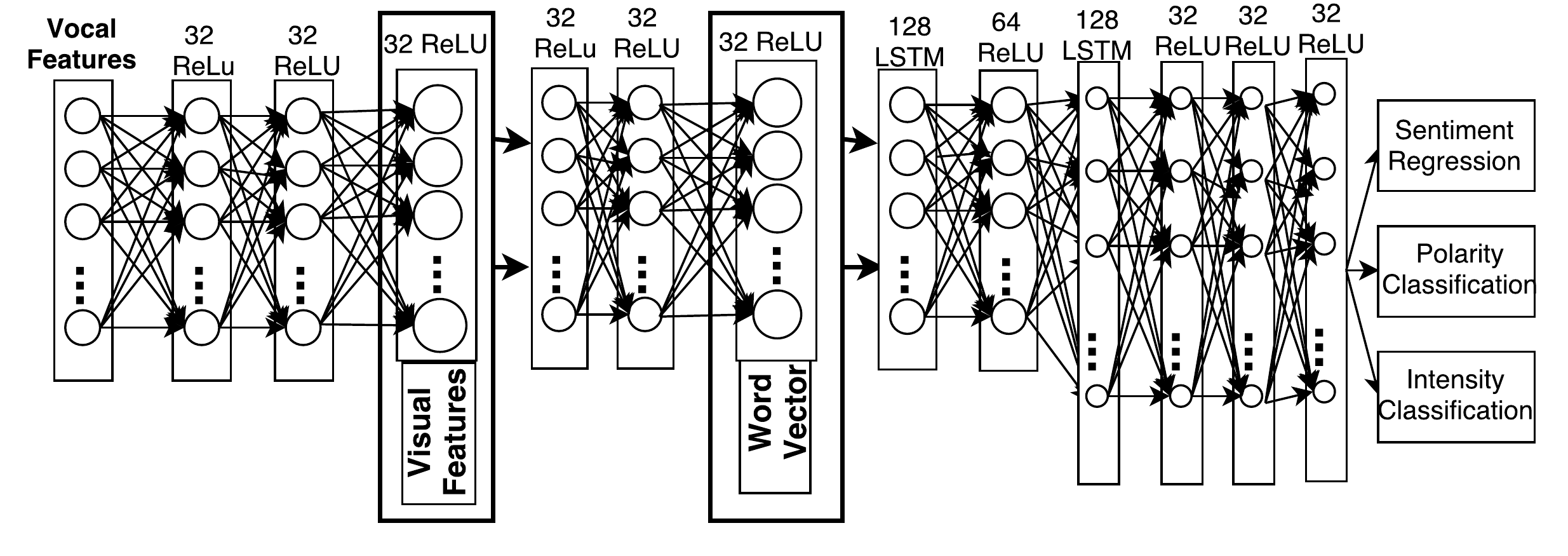}
\caption{Structure of HF tri-task model}\label{fig_HL}
\end{figure}

The structure of the EF model is shown in Figure~\ref{fig_FL}. The feature vectors are simply concatenated in the EF model. A drop-out rate of 0.2 is applied to the combined feature vector. We then stack one layer of 128 LSTM units and three layers of 32 ReLU units with an L2 regularizer weight of 0.01 on top of the multimodal inputs. To compare performance of the fusion strategies, this same structure is applied to the multimodal inputs in all multimodal models. In the TFN model, we compute the Cartesian products (shown in Figure~\ref{fig_TFN}) of the unimodal model top layers as the multimodal inputs. Unlike \citet{zadeh2017tensor}, we did not add the extra constant dimension with value 1 when computing the 3-fold Cartesian space in order to reduce the dimensionality of the multimodal input. In the LF model, as shown in Figure~\ref{fig_DL}, we concatenate the unimodal model top layers as the multimodal inputs. In the HF model, unimodal information is used in a hierarchy where the top layer of the lower unimodal model is concatenated with the input layer of the higher unimodal model, as shown in Figure~\ref{fig_HL}. We use the vocal modality at the bottom of the hierarchy while using the verbal modality at the top in HF fusion. This is because in previous studies (e.g.,~\citet{zadeh2018multi}) the verbal modality was shown to be the most effective for unimodal sentiment analysis, while the vocal modality was shown to be the least effective.

\section{Experiments and Results}\label{sec_ex}
Here we report our sentiment score prediction experiments.\footnote{Source code available at: \url{https://github.com/tianleimin/}.} In Tables~\ref{result_uni} and \ref{result_multi}, ``S'' is the single-task learning model; ``S+P'' is the bi-task learning model with polarity classification as the auxillary task; ``S+I'' is the bi-task learning model with intensity classification as the auxillary task; ``S+P+I'' is the tri-task learning model. To evaluate the performance of sentiment score prediction, following previous work \citep{zadeh2018multi}, we report both Pearson's correlation coefficient (CC, higher is better) and mean absolute error (MAE, lower is better) between predictions and annotations of sentiment scores on the test set. In each row of Tables~\ref{result_uni} and \ref{result_multi}, the numbers in bold are the best performance for each modality or fusion strategy. To identify the significant differences in results, we perform a two-sample Wilcoxon test on the sentiment score predictions given by each pair of models being compared and consider $p < 0.05$ as significant. We also include random prediction as a baseline and the human performance reported by \citet{zadeh2017tensor}.

\subsection{Unimodal Experiments}\label{ex_uni}
The results of unimodal sentiment prediction experiments are shown in Table~\ref{result_uni}.\footnote{All unimodal models have significantly different performance. $p = 0.009$ for S+P and S+P+I Visual models, $p << 0.001$ for Visual and Vocal S+I models.} The verbal models have the best performance here, which is consistent with previous sentiment analysis studies on multiple databases (e.g.,~\citet{zadeh2018multi}). This suggests that lexical information remains the most effective for sentiment analysis. On each modality, the best performance is achieved by a multi-task learning model. This answers our first research question and suggests that sentiment analysis can benefit from multi-task learning. 

In multi-task learning, the main task gains additional information from the auxillary tasks. Compared to the S model, the S+P model has increased focus on the polarity of sentiment, while the S+I model has increased focus on the intensity of sentiment. On the verbal modality, the S+P model achieved the best performance, while on the visual modality the S+I model achieved the best performance. This suggests that the verbal modality is weaker at communicating the polarity of sentiment. Thus, verbal sentiment analysis benefits more from including additional information on polarity. On the contrary, the visual modality is weaker at communicating the intensity of sentiment. Thus, visual sentiment analysis benefits more from including additional information on intensity. For the vocal modality, the S+P+I model achieved the best performance, and the S+P model yielded improved performance over that of the S model. This suggests that the vocal modality is weaker at communicating the polarity of sentiment. Thus, addressing our second research question, the results suggest that individual modalities differ when conveying each aspect of sentiment.

\begin{table}[!htb]
\begin{center}
\begin{tabular}{|l|cccc|}
\hline
\textbf{CC} & \textbf{S} & \textbf{S+P} & \textbf{S+I} & \textbf{S+P+I} \\
\hline
Random & -- & -- & -- & -- \\
Vocal  & 0.125 & 0.149 & 0.119 & \textbf{0.153} \\
Visual & 0.092 & 0.109 & \textbf{0.116} & 0.106 \\
Verbal & 0.404 & \textbf{0.455} & 0.434 & 0.417 \\
Human & 0.820 & -- & -- & -- \\
\hline
\textbf{MAE} & \textbf{S} & \textbf{S+P} & \textbf{S+I} & \textbf{S+P+I} \\
\hline
Random & 1.880 & -- & -- & -- \\
Vocal  & 1.456 & 1.471 & 1.444 & \textbf{1.431} \\
Visual & 1.442 & \textbf{1.439} & 1.453 & 1.460 \\
Verbal & 1.196 & \textbf{1.156} & 1.181 & 1.206 \\
Human & 0.710 & -- & -- & -- \\
\hline
\end{tabular}
\end{center}
\caption{Unimodal sentiment analysis results on the CMU-MOSI test set. Numbers in bold are the best results on each modality.}
\label{result_uni}
\end{table}


\subsection{Multimodal Experiments}\label{ex_multi}
The results of the multimodal experiments are shown in Table~\ref{result_multi}. We find that EF$>$HF$>$TFN$>$LF.\footnote{Performance of multimodal models are significantly different, except that the HF S and the TFN S+P model have $p = 0.287$. $p = 0.001$ for EF S+P+I and HF S, $p = 0.017$ for TFN S+P and LF S.} The reason that the EF model yields the best performance may be that it is the least complex. This is shown to be beneficial for the small CMU-MOSI database \citep{poria2015deep}. Unlike \citet{zadeh2017tensor}, here the EF model outperforms the TFN model. However, the TFN model achieved the best performance on the training and validation sets. This indicates that performance of the TFN model may be limited by over-fitting. Compared to the feature concatenation used in EF, the Cartesian product used in TFN results in higher dimensionality of the multimodal input vector,\footnote{Dimension of the EF input is 420, for TFN is 65,536.} which in turn increases the complexity of the model. Similarly, the HF model has worse performance than the EF model here, unlike in \citet{tian2016recognizing}. This may be due to the HF model having the deepest structure with the most hidden layers, which increases its complexity.

The performance of unimodal and multimodal models are significantly different. In general, the multimodal models have better performance than the unimodal models.\footnote{Except that the LF models often have worse performance than the verbal S+P model. $p << 0.001$ for TFN S+P and verbal S+P, $p = 0.017$ for verbal S+P and LF S.} Unlike unimodal models, multimodal models benefit less from multi-task learning. In fact, the HF and LF models have better performance using single-task learning. For the TFN models, only the S+P model outperforms the S model, although the improvement is not significant.\footnote{$p = 0.105$ for S TFN and S+P TFN.} For the EF models, multi-task learning results in better performance.\footnote{$p = 0.888$ for S EF and S+P EF, $p = 0.029$ for S EF and S+I EF, $p = 0.009$ for S EF and S+P+I EF.} The reason that EF benefits from multi-task learning may be that it combines modalities without bias and individual features have more influence on the EF model. Thus, the benefit of multi-task learning is preserved in EF. However, the other fusion strategies (TFN, LF, HF) attempt to compensate one modality with information from other modalities, i.e., relying more on other modalities when one modality is weaker at predicting an aspect of sentiment. In Section~\ref{ex_uni} we showed that each modality has different weaknesses when conveying the polarity or intensity aspect of sentiment. The multimodal models are able to overcome such weaknesses by modality fusion. Thus, multi-task learning does not yield additional improvement in these models. Our observations answer our third research question: multi-task learning influences unimodal and multimodal sentiment analysis differently.

\begin{table}[!htb]
\begin{center}
\begin{tabular}{|l|cccc|}
\hline
\textbf{CC} & \textbf{S} & \textbf{S+P} & \textbf{S+I} & \textbf{S+P+I} \\
\hline
Random & -- & -- & -- & -- \\
EF & 0.471 & 0.472 & 0.476 & \textbf{0.482} \\
TFN & 0.448 & \textbf{0.461} & 0.446 & 0.429 \\
LF & \textbf{0.454} & 0.413 & 0.428 & 0.428 \\
HF & \textbf{0.469} & 0.424 & 0.458 & 0.432 \\
Human & 0.820 & -- & -- & -- \\
\hline
\textbf{MAE} & \textbf{S} & \textbf{S+P} & \textbf{S+I} & \textbf{S+P+I} \\
\hline
Random & 1.880 & -- & -- & -- \\
EF & 1.197 & 1.181 & 1.193 & \textbf{1.172} \\
TFN & 1.186 & 1.181 & \textbf{1.178} & 1.205 \\
LF & \textbf{1.179} & 1.211 & 1.204 & 1.201 \\
HF & \textbf{1.155} & 1.211 & 1.164 & 1.187 \\
Human & 0.710 & -- & -- & -- \\
\hline
\end{tabular}
\end{center}
\caption{Multimodal sentiment analysis results on the CMU-MOSI test set. Numbers in bold are the best results for each fusion strategy in each row.}
\label{result_multi}
\end{table}


\section{Discussion}\label{sec_diss}
Our unimodal experiments in Section~\ref{ex_uni} show that unimodal sentiment analysis benefits significantly from multi-task learning. As suggested by \citet{wilson2008fine}, polarity and intensity can be conveyed through different units of language. We can use one word such as \textit{extremely} to express intensity, while the polarity of a word and the polarity of the opinion segment the word is in may be opposite. Our work supports a fine-grained sentiment analysis. By including polarity and intensity classification as the auxiliary tasks, we illustrate that individual modalities differ when conveying sentiment. In particular, the visual modality is weaker at conveying the intensity aspect of sentiment, while the vocal and verbal modalities are weaker at conveying the polarity aspect of sentiment. In previous emotion recognition studies under the circumplex model of emotions \citep{russell1980circumplex}, it was found that the visual modality is typically weaker at conveying the Arousal dimension of emotion, while the vocal modality is typically weaker at conveying the Valence dimension of emotion (e.g.,~\citet{nicolaou2011continuous}). The similarities between the performance of different communication modalities on conveying emotion dimensions and on conveying different aspects of sentiment indicate a connection between emotion dimensions and sentiment. The different behaviors of unimodal models in conveying the polarity and intensity aspects of sentiment also explain the improved performance achieved by modality fusion in Section~\ref{ex_multi} and in various previous studies. By decomposing sentiment scores into polarity and intensity, our work provides detailed understanding on how individual modalities and multimodal information convey these two aspects of sentiment.

We are aware that performance of our sentiment analysis models leaves room for improvement compared to state-of-the-art on the CMU-MOSI database. One reason may be that we did not perform pre-training in this study. In the future, we plan to explore more advanced learning techniques and models, such as a Dynamic Fusion Graph \citep{cmumoseiacl2018}, to improve performance. We also plan to perform case studies to provide detailed analysis on how the unimodal models benefit from multi-task learning, and how individual modalities compensate each other in the multimodal models.

\section{Conclusions}\label{sec_conc}
In this work, we decouple Likert scale sentiment scores into two aspects: polarity and intensity, and study the influence of including polarity and/or intensity classification as auxiliary tasks to sentiment score regression. Our experiments showed that all unimodal models and some multimodal models benefit from multi-task learning. Our unimodal experiments indicated that each modality conveys different aspects of sentiment differently. In addition, we observed similar behaviors between how individual modalities convey the polarity and intensity aspects of sentiments and how they convey the Valence and Arousal emotion dimensions. Such connections between sentiments and emotions encourage researchers to obtain an integrated view of sentiment analysis and emotion recognition. Our multimodal experiments showed that unlike unimodal models, multimodal models benefit less from multi-task learning. This suggests that one reason that modality fusion yields improved performance in sentiment analysis is its ability to combine the different strengths of individual modalities on conveying sentiments.

Note that we only conducted experiments on the CMU-MOSI database. In the future, we plan to expand our study to multiple databases. Moreover, we are interested in including databases collected on modalities beyond the three Vs. For example, gestures or physiological signals. We also plan to perform sentiment analysis and emotion recognition in a multi-task learning setting to further explore the relationship between sentiments and emotions.

\section*{Acknowledgments}
We would like to thank Zack Hodari for his support on computational resources, and Jennifer Williams for the insightful discussion.

\bibliography{acl2018}
\bibliographystyle{acl_natbib}

\end{document}